\documentclass[11pt,a4paper]{article}
\usepackage[hyperref]{naaclhlt2018}
\usepackage{times}
\usepackage{latexsym}

\usepackage{xr}
\usepackage{bm}
\usepackage{enumitem}
\usepackage[hang,flushmargin]{footmisc}
\usepackage{booktabs}
\usepackage{multirow}
\usepackage{graphicx}
\usepackage{amssymb}
\usepackage{amsmath}
\usepackage{verbatim}
\usepackage{breakcites}
\usepackage{pifont}
\newcommand{\cmark}{\text{\ding{51}}}
\newcommand{\xmark}{\text{\ding{55}}}
\usepackage[hyphenbreaks]{breakurl}
\usepackage{url}

\aclfinalcopy 
\def\aclpaperid{779} 

\title{Modeling Semantic Plausibility by Injecting World Knowledge}

\author{
Su Wang$^{1,2}$\ \ \ Greg Durrett$^{3}$\ \ \ Katrin Erk$^{1}$
\vspace{0.5ex} \\
$^{1}$Department of Linguistics \\
$^{2}$Department of Statistics and Data Science \\
$^{3}$Department of Computer Science 
\vspace{0.5ex} \\
The University of Texas at Austin \\
{\tt \small{shrekwang@utexas.edu}\ \  \small{gdurrett@cs.utexas.edu}\ \  \small{katrin.erk@mail.utexas.edu}}
}

\begin{document}

\maketitle

\vspace{-2ex}
\begin{abstract}
Distributional data tells us that a man can swallow candy, but not that a man can swallow a paintball, since this is never attested. However both are physically plausible events. This paper introduces the task of semantic plausibility: recognizing plausible but possibly novel events. We present a new crowdsourced dataset of semantic plausibility judgments of single events such as \emph{man swallow paintball}. Simple models based on distributional representations perform poorly on this task, despite doing well on selection preference, 
but injecting manually elicited knowledge about entity properties provides a substantial performance boost. Our error analysis shows that our new dataset is a great testbed for semantic plausibility models: more sophisticated knowledge representation and propagation could address many of the remaining errors.
\end{abstract}
\smallskip

\section{Introduction}
\label{sec:intro}

\externaldocument{02-related-work}
\externaldocument{03-data}
\externaldocument{04-feat}
\externaldocument{05-models}
\externaldocument{06-exp-rets}
\externaldocument{07-error}
\externaldocument{08-conclusion}

Intuitively, a \textit{man} can \textit{swallow} a \textit{candy} or \textit{paintball} but not a \textit{desk}. Equally so, one cannot plausibly \textit{eat} a \textit{cake} and then \textit{hold} it. What kinds of semantic knowledge are necessary for distinguishing a physically plausible event (or event sequence) from an implausible one? \emph{Semantic plausibility} stands in stark contrast to the familiar \emph{selectional preference} \cite{Erk:10,Cruys:14} which is concerned with the \emph{typicality} of events (Table 1). For example, \textit{candy} is a typical entity for \textit{man-swallow-*} but \textit{paintball} is not, even though both events are plausible physically. Also, some events are physically plausible but are never stated because humans avoid stating the obvious. Critically, semantic plausibility is sensitive to certain properties such as relative object size that are not explicitly encoded by selectional preferences \cite{Bagherinezhad:16}. Therefore, it is crucial that we learn to model these dimensions in addition to using classical distributional signals.




\begin{table}[!t]
\begin{center}
\begin{tabular}{lccc}
\toprule
{\small \textit{man-swallow-*}}& \small{\textsc{preferred?}} & \small{\textsc{plausible?}} \\
\midrule
{\small \textit{-candy}} & \cmark & \cmark \\
{\small \textit{-paintball}} & \xmark & \cmark \\
{\small \textit{-desk}} & \xmark & \xmark \\
\bottomrule
\end{tabular}
\end{center}
\caption{Distinguishing \emph{semantic plausibility} from \emph{selectional preference}. \textit{candy} is selectionally preferred because it is distributionally common patient in the event \textit{man-swallow-*}, as opposed to the bizarre and rarely seen (if at all) patient \textit{paintball}. However both are semantically plausible according to our world knowledge: they are small-sized objects that are swallowable by a man. \textit{desk} is both distributionally unlikely and implausible (i.e. oversized for swallowing).}

\end{table}


\medskip

\noindent
Semantic plausibility is pertinent and crucial in a multitude of interesting NLP tasks put forth previously, such as narrative schema \cite{Chambers:13}, narrative interpolation \cite{Bowman:16}, story understanding \cite{mostafazadeh-EtAl:2016:N16-1}, and paragraph reconstruction \cite{Li:17}. Existing methods for these tasks, however, draw predominantly (if not only) on distributional data and produce rather weak performance. Semantic plausibility over \emph{subject-verb-object} triples, while simpler than these other tasks, is a key building block that requires many of the same signals and encapsulates complex world knowledge in a binary prediction problem.


In this work, we show that world knowledge injection is necessary and effective for the semantic plausibility task, for which we create a robust, high-agreement dataset (details in section 3). Employing methods inspired by the recent work on world knowledge propagation through distributional context \cite{Forbes:17,Wang:17}, we accomplish the goal with minimal effort in manual annotation. Finally, we perform an in-depth error analysis to point to future directions of work on semantic plausibility.

\section{Related Work}
\label{sec:related-work}

Simple events (i.e. \textsc{s-v-o}) have seen thorough investigation from the angle of selectional preference. While early works are resource-based \cite{Resnik:1996,Clark:01}, later work shows that unsupervised learning with distributional data yields strong performance \cite{O-Seaghdha:10,Erk:10}, which has recently been further improved upon with neural approaches \cite{Cruys:14,Tilk:16}. Distribution-only models however, as will be shown, fail on the semantic plausibility task we propose.

Physical world knowledge modeling appears frequently in more closely related work. \citet{Bagherinezhad:16} combine computer vision and text-based information extraction to learn the relative sizes of objects; \citet{Forbes:17} crowdsource physical knowledge along specified dimensions and employ belief propagation to learn relative physical attributes of object pairs. \citet{Wang:17} propose a multimodal LDA to learn the definitional properties (e.g. \textit{animal, four-legged}) of entities. \citet{Zhang:17} study the role of common-sense knowledge in natural language inference, which is inherently between-events rather than single-event focused. Prior work does not specifically handles the (single-event) semantic plausibility task and related efforts do not necessarily adapt well to this task, as we will show, suggesting that new approaches are needed.


\section{Data}
\label{sec:data}

\externaldocument{02-related-work}
\externaldocument{03-data}
\externaldocument{04-feat}
\externaldocument{05-models}
\externaldocument{06-exp-rets}
\externaldocument{07-error}
\externaldocument{08-conclusion}

To study the semantic plausibility of \textsc{s-v-o} events, specifically \emph{physical} semantic plausibility, we create a dataset\footnote{Link: \url{https://github.com/suwangcompling/Modeling-Semantic-Plausibility-NAACL18/tree/master/data}.} through Amazon Mechanical Turk with the following criteria in mind: (i) \emph{Robustness}: Strong inter-annotator agreement; (ii) \emph{Diversity}: A wide range of typical/atypical, plausible/implausible events; (iii) \emph{Balanced}: Equal number of plausible and implausible events. 

In creating physical events, we work with a fixed vocabulary of 150 concrete verbs and 450 concrete nouns from \citet{Brysbaert:14}'s word list, with a concreteness threshold of 4.95 (scale: 0-5). We take the following steps:
\begin{enumerate}[label=(\alph*),topsep=1.5pt,itemsep=-1ex,partopsep=1.5ex,parsep=1.5ex]
\item Have Turkers write down plausible or implausible \textsc{s-v} and \textsc{v-o} selections;
\item Randomly generate \textsc{s-v-o} triples from collected \textsc{s-v} and \textsc{v-o} pairs;
\item Send resulting \textsc{s-v-o} triples to Turkers to filter for ones with high agreement (by majority vote).
\end{enumerate}
(a) ensures diversity and the cleanness of data (compared with noisy selectional preference data collected unsupervised from free text): the Turkers are instructed (with examples) to (i) consider both typical and atypical selections (e.g. \textit{man-swallow-*} with \textit{candy} or \textit{paintball}); (ii) disregard metaphorical uses (e.g. \textit{feel-blue} or \textit{fish-idea}). 2,000 pairs are collected in the step, balancing typical and atypical pairs. In (b), we manually filter error submissions in triple generation. For (c), 5 Turkers provide labels, and we only keep the ones that have $\geq 3$ majority votes, resulting with 3,062 triples (of 4,000 annotated triples, plausible-implausible balanced), with \textbf{100\% $\bm{\geq 3}$ agreement, 95\% $\bm{\geq 4}$ agreement, and 90\% $\bm{5}$ agreement}. 
\medskip

\noindent
To empirically show the failure of distribution-only methods, we run \citet{Cruys:14}'s neural net classifier (hereforth \textbf{\textsc{nn}}), which is one of the strongest models designed for selectional preference (Figure 1, left-box). Let $\bm{x}$ be the concatenation of the embeddings of the three words in an \textsc{s-v-o} triple. The prediction $P(y|\bm{x})$ is computed as follows:
\begin{equation}
P(y=1|\bm{x}) = \sigma_2(W_2 \sigma_1(W_1\bm{x}))
\end{equation}
where $\sigma$ is a nonlinearity, $W$ are weights, and we use 300D pretrained GloVe vectors \cite{Pennington:14}. The model achieves an accuracy of 68\% (logistic regression baseline: 64\%) after fine-tuning, verifying the intuition that distributional data alone cannot satisfactorily capture the semantics of physical plausibility.

\section{World Knowledge Features}
\label{sec:feat}

\externaldocument{02-related-work}
\externaldocument{03-data}
\externaldocument{04-feat}
\externaldocument{05-models}
\externaldocument{06-exp-rets}
\externaldocument{07-error}
\externaldocument{08-conclusion}

Recognizing that a distribution-alone method lacks necessary information, we collect a set of world knowledge features. The feature types derive from inspecting the high agreement event triples for knowledge missing in distributional selection (e.g. relative sizes in \textit{man-swallow-paintball/desk}). Previously, \citet{Forbes:17} proposed a three level (\textsc{3-level}) featurization scheme, where an object-pair can take 3 values for, e.g. relative size: $\{-1,0,1\}$ (i.e. lesser, similar, greater). This method, however, does not explain many cases we observed. For instance, \textit{man-hug-cat/ant}, \textit{man} is larger than both \textit{cat} and \textit{ant}, but the latter event is implausible. \textsc{3-level} is also inefficient: $k$ objects incur $O(k^2)$ elicitations. We thus propose a binning-by-landmark method, which is sufficiently fine-grained while still being efficient and easy for the annotator: given an entity $n$, the Turker decides to which of the landmarks $n$ is closest to. E.g., for \textsc{size}, we have the landmarks \textit{\{watch, book, cat, person, jeep, stadium\}}, in ascending sizes. If $n =$ \textit{dog}, the Turker may put $n$ in the bin corresponding to \textit{cat}. The features\footnote{We experimented with numerous feature types, e.g. size, temperature, shape, etc. and kept the subset that contributes most substantially to semantic plausibility classification. More details on the feature types in supplementary material (\url{https://github.com/suwangcompling/Modeling-Semantic-Plausibility-NAACL18/tree/master/supplementary}).} are listed with their landmarks as follows:
\begin{enumerate}[label=$\bullet$,topsep=1.5pt,itemsep=-1ex,partopsep=1.5ex,parsep=1.5ex]
\item {\small \textsc{sentience}: \textit{rock, tree, ant, cat, chimp, man}}.
\item {\small \textsc{mass-count}: \textit{milk, sand, pebbles, car}}.
\item {\small \textsc{phase}:  \textit{smoke, milk, wood}}.
\item {\small \textsc{size}: \textit{watch, book, cat, person, jeep, stadium}}.
\item {\small \textsc{weight}: \textit{watch, book, dumbbell, man, jeep, stadium}}.
\item {\small \textsc{rigidity}: \textit{water, skin, leather, wood, metal}}. 
\end{enumerate}
5 Turkers provide annotations for all 450 nouns, and we obtained \textbf{93\% $\bm{\geq 3}$ agreement, 85\% $\bm{\geq 4}$ agreement, and 79\% 5 agreement}. 

Our binning is sufficiently granular, which is crucial for semantic plausibility of an event in many cases. E.g. for \textit{man-hug-cat/ant}, \textit{man, cat} and \textit{ant} fall in the $4^{th}, 3^{rd}$ and $1^{st}$ bin, which suffices to explain why \textit{man-hug-cat} is plausible while \textit{man-hug-ant} is not. Compared to past work \cite{Forbes:17}, it is efficient. Each entity only needs one assignment in comparison to the landmarks to be located in a ``global scale'' (e.g. from the smallest to the largest objects), and even for extreme granularity, it only takes $O(k \log k)$ comparisons. It is also intuitive: differences in bins capture the intuition that one can \textit{hug} smaller objects as long as those objects are not too small.

\section{Models}
\label{sec:models}

We answer two questions: (i) Does world knowledge improve the accuracy of semantic plausibility classification? (ii) Can we minimize effort in knowledge feature annotation by learning from a small amount of training data? 

\begin{figure}[t]
\begin{center}
\scalebox{0.95}{
\includegraphics[width=\columnwidth]{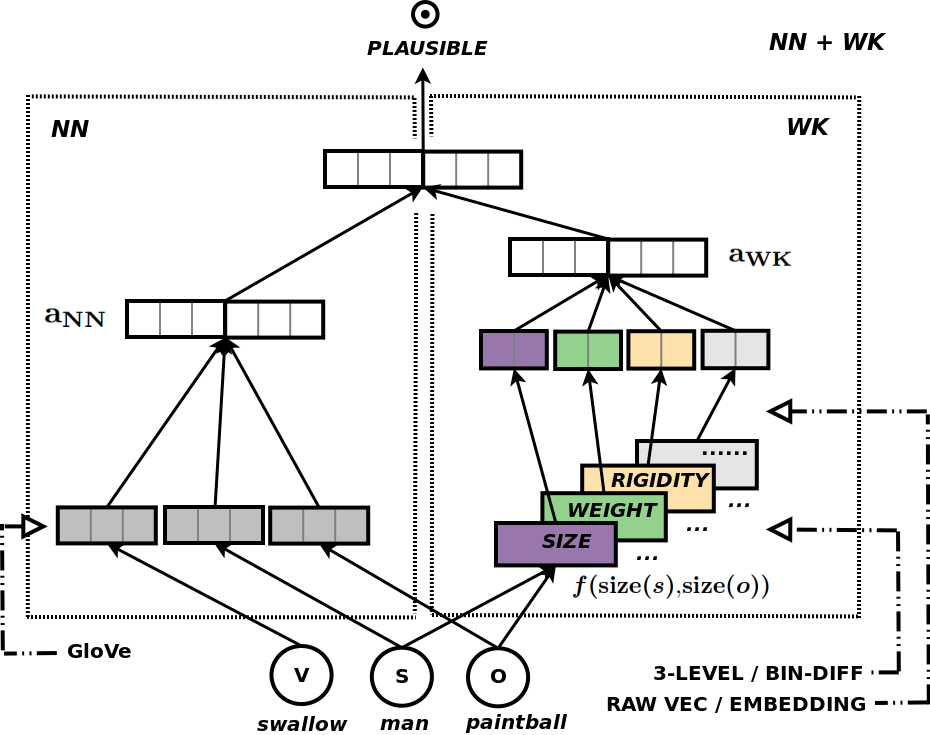}}
\end{center}
\label{fig:nn-wk}
\caption{Model architecture (example input: \textit{man-swallow-paintball}). Left box: \citet{Cruys:14}'s neural net (\textbf{\textsc{nn}}, embeddings only); Right box: world knowledge feature net \textbf{\textsc{wk}} with different modeling choices (Section \ref{sec:models}). Only \textsc{size, weight, rigidity} are shown; the rest receive the same treatment. \textbf{\textsc{nn + wk}}: embedding and world knowledge combined.}
\end{figure}

For question (i), we experiment with various methods to incorporate the features on top of the embedding-only \textbf{\textsc{nn}} (Section \ref{sec:data}). Our architecture\footnote{More configuration details in supplementary material.} is outlined in Figure 1, where we ensemble the \textbf{\textsc{nn}} (left-box) and another feedforward net for features (\textbf{\textsc{wk}}, right-box) to produce the final prediction. For the feature net, the relative physical attributes of the subject-object pair can be encoded in \textbf{\textsc{3-level}} (Section \ref{sec:feat}) or the bin difference (\textbf{\textsc{bin-diff}}) scheme.\footnote{We also tried using \emph{bin numbers} directly, however it does not produce ideal results (classification accuracy between \textsc{3-level} and \textsc{bin-diff}). Thus for brevity we drop this setup.} For \textsc{bin-diff}, given the two entities in an \textsc{s-v-o} event (i.e. \textsc{s, o}) \textit{ant} and \textit{man}, which are in the bins of the landmark \textit{watch} (i.e. the $1^{st}$) and that of \textit{person} (i.e. the $4^{th}$), the pair \textit{ant-man} gets a \textsc{bin-diff} value of $1-4=-3$.  Exemplifying the featurization function $f(s,o)$ with \textsc{size}:
\begin{align}
f_{\textsc{3-L}}(\textsc{size}(s),\textsc{size}(o)) &\in \{-1,0,1\} \\
f_{\textsc{bin}}(\textsc{size}(s),\textsc{size}(o)) &= \textsc{bin}(s)-\textsc{bin}(o)
\end{align}
Then, given a featurization scheme, we may feed raw feature values (\textbf{\textsc{raw vec}}, for \textsc{3-level}, e.g. concatenation of -1, 0 or 1 of all feature types, in that order, and in one-hot format), or feature embeddings (\textbf{\textsc{embedding}}, e.g. concatenation of embeddings looked up with feature values). Finally, let $\bm{a}_{\textbf{\textsc{nn}}},\bm{a}_{\textbf{\textsc{wk}}}$ be the penultimate-layer vectors of \textbf{\textsc{nn}} and \textbf{\textsc{wk}} (see Figure 1), we affine transform their concatenation to predict label $\hat{y}$ with argmax on the final softmax layer:
\begin{equation}
\hat{y} = \underset{y}{\text{argmax}}\ \text{softmax}(\sigma(W[\bm{a}_{\textbf{\textsc{nn}}};\bm{a}_{\textbf{\textsc{wk}}}] + \bm{b}))
\end{equation}
where $\sigma$ is a ReLU nonlinearity. We will only report the results from the best-performing model configuration, which has \textsc{bin-diff} + \textsc{embedding}. The model will be listed below as \textbf{\textsc{nn + wk-gold}} (i.e. with \textbf{\textsc{gold}}, Turker-annotated \textbf{\textsc{w}}orld \textbf{\textsc{k}}nowledge features).

\begin{table}[!t]
\begin{center}
\scalebox{0.7}{
\begin{tabular}{lcccc}
\toprule
\textsc{models} & \multicolumn{2}{c}{5\%} & \multicolumn{2}{c}{20\%} \\
\midrule
Label Spreading \citep{Zhou:04} & \multicolumn{2}{c}{0.56}  & \multicolumn{2}{c}{0.59}  \\
Factor Graph \citep{Forbes:17} & \multicolumn{2}{c}{0.69} & \multicolumn{2}{c}{0.71} \\
Multi-LDA \citep{Wang:17} & \multicolumn{2}{c}{0.64} & \multicolumn{2}{c}{0.72} \\
\midrule
Logistic Regression & \multicolumn{2}{c}{0.72} & \multicolumn{2}{c}{0.83} \\
Factor Graph (initialized with our LR) & \multicolumn{2}{c}{0.72} & \multicolumn{2}{c}{0.84} \\
Ordinal-LR & \multicolumn{2}{c}{\textbf{0.76}} & \multicolumn{2}{c}{\textbf{0.88}} \\
\midrule
\midrule
\multirow{2}{*}{\textsc{models}} & \multicolumn{2}{c}{5\%} & \multicolumn{2}{c}{20\%} \\
\cmidrule{2-5}
& \textsc{3-l} & \textsc{bin} & \textsc{3-l} & \textsc{bin} \\
\midrule
Logistic Regression & 0.61 & 0.21 & 0.68 & 0.26 \\
Ordinal-LR & \textbf{0.66} & \textbf{0.32} & \textbf{0.76} & \textbf{0.40} \\
\bottomrule
\end{tabular}}
\end{center}
\label{tab:feat-prop}
\caption{Feature Propagation. Top-table: results on \citet{Forbes:17}'s 2.5k object pair data; Bottom-table: results on our 10k object pair data.}
\end{table}

For question (ii), we select a data-efficient feature learning model. Following \citet{Forbes:17} we evaluate the models with 5\% or 20\% of training data. We experiment with several previously proposed techniques: (a) \emph{label spreading}; (b) \emph{factor graph}; (c) \emph{multi-LDA}. As a baseline we employ a simple but well-tuned logistic regressor (LR). We also initialize the factor graph with this LR, on account of its unexpectedly strong performance.\footnote{We verified our setup with the authors and they attributed the higher performance of our LR to hyperparameter choices.} Finally, observing that the feature types are inherently ordinal (e.g. \textsc{size} from small to large), we also run \emph{ordinal logistic regression} \cite{Adeleke:10}. For model selection we first evaluate the object-pair attribute data collected by \citet{Forbes:17}, 2.5k pairs labeled in the \textsc{3-level} scheme. We then compared the the LR and Ordinal-LR (our strongest models\footnote{Because the factor graph + LR gives very slight improvement, for simplicity we choose LR instead.} in this experiment) on 10k randomly generated object-pairs from our annotated nouns. The results are summarized in Table 2, where we see (i) \textsc{3-level} propagation is much easier; (ii) our object-pairs are more challenging, likely due to sparsity with larger vocabulary size; (iii) ordinality information contributes substantially to performance. The model that uses propagated features (w/ Ordinal-LR) will be listed as \textbf{\textsc{nn + wk-prop}}.

\section{Semantic Plausibility Results}
\label{sec:exp-rets}

\externaldocument{02-related-work}
\externaldocument{03-data}
\externaldocument{04-feat-prop}
\externaldocument{05-models}
\externaldocument{06-exp-rets}
\externaldocument{07-error}
\externaldocument{08-conclusion}

We evaluate the models on the task of classifying our 3,062 \textsc{s-v-o} triples by semantic plausibility (10-fold CV, taking the average over 20 runs with the same random seed). 
We compare our three models in the \textsc{3-level} and \textsc{bin-diff} schemes, with \textbf{\textsc{nn + wk-prop}} evaluated in 5\% and 20\% training conditions. The results are outlined in Table 3.
Summarizing our findings: (i) world knowledge undoubtedly leads to a strong performance boost ($\sim$8\%); (ii) \textsc{bin-diff} scheme works much better than \textsc{3-level} --- it manages to outperform the latter even with much weaker propagation accuracy; (iii) the accuracy loss with propagated features seems rather mild with 20\% labeled training and the best  scheme.  

\begin{table}[!t]
\begin{center}
\scalebox{0.7}{
\begin{tabular}{lccccc}
\toprule
\textsc{models} & \multicolumn{4}{c}{\textsc{accuracy}} \\
\midrule
Random & \multicolumn{4}{c}{0.50} \\
LR baseline & \multicolumn{4}{c}{0.64} \\
\textbf{\textsc{nn}} \citep{Cruys:14} & \multicolumn{4}{c}{0.68} \\
\textbf{\textsc{nn + wk-gold}} & \multicolumn{4}{c}{\textbf{0.76}} \\
\midrule
\multirow{3}{*}{\textbf{\textsc{nn + wk-prop}}} &  \multicolumn{2}{c}{5\%} & \multicolumn{2}{c}{20\%} \\
\cmidrule{2-5}
& \textsc{3-l} & \textsc{bin} & \textsc{3-l} & \textsc{bin} \\
\cmidrule{2-5}
& 0.69 & 0.70 & 0.71 & 0.74 \\
\bottomrule
\end{tabular}}
\end{center}
\label{fig2:results}
\caption{Semantic Plausibility (binary) Classification. The average of 10-fold CV (splitting on the total 3,062 entries). The neural classifier injected with full annotation of world knowledge (i.e. \textbf{\textsc{nn + wk-gold}}) performs substantially better, and the performance retainment is rather strong with propagated features (by Ordinal-LR) from small fractions of gold annotation (i.e. in \textbf{\textsc{nn + wk-prop}}).}
\end{table}

\section{Error Analysis}
\label{sec:error}

\externaldocument{02-related-work}
\externaldocument{03-data}
\externaldocument{04-feat-prop}
\externaldocument{05-models}
\externaldocument{06-exp-rets}
\externaldocument{07-error}
\externaldocument{08-conclusion}

To understand what challenges remain in this task, we run the models above 200 times (10-fold CV, random shuffle at each run), and inspect the top 200 most frequently misclassified cases. The percentage statistics below are from counting the error cases. 
\medskip

\noindent
In the cases where \textbf{\textsc{nn}} misclassifies while \textbf{\textsc{nn + wk-gold}} correctly classifies, 60\% relates to \textsc{size} and \textsc{weight} (e.g. missing \textit{man-hug-ant} (bad) or \textit{dog-pull-paper} (good)). \textsc{phase} takes up 18\% (e.g. missing \textit{monkey-puff-smoke} (good)). This validates the intuition that distributional contexts do not encode these types of world knowledge.
\medskip

\noindent
For cases often misclassified by \emph{all} the models, we observe two main types of errors: (i) data sparsity; (ii) highly-specific attributes. 
\medskip

\noindent
\textbf{Data sparsity (32\%)}. \textit{man-choke-ant}, e.g., is a singleton big-object-choke-small-object instance, and there are no distributionally similar verbs that can help (e.g. \textit{suffocate}); For \textit{sun-heat-water}, because the majority of the actions in the data are limited to solid objects, the models tend to predict implausible for whenever a gas/liquid appears as the object.
\medskip

\noindent
\textbf{Highly-specific attributes (68\%)}. ``long-tailed'' physical attributes which are absent from our feature set are required. To exemplify a few:\footnote{Percentages calculated with the 68\% as the denominator. Full list in supplementary material.}
\begin{enumerate}[label=$\bullet$,topsep=1.5pt,itemsep=-1ex,partopsep=1.5ex,parsep=1.5ex]
\item \emph{edibility} (21\%). \textit{*-fry-egg} (plausible) and \textit{*-fry-cup} (implausible) are hard to distinguish because \textit{egg} and \textit{cup} are similar in \textsc{size}/\textsc{weight}/..., however introducing large free-text data to help learn edibility misguides our model to mind selectional preference, causing mislabeling of other events.
\item \emph{natural vs. artificial} (18\%). Turkers often think creating natural objects like \textit{moon} or \textit{mountain} is implausible but creating an equally big (but artificial) object like \textit{skyscraper} is plausible.
\item \emph{hollow objects} (15\%). \textit{plane-contain-shell} and \textit{purse-contain-scissors} are plausible, but the hollow-object-can-contain-things attribute is failed to be captured.
\item \emph{forefoot dexterity} (5\%). \textit{horse-hug-man} is implausible but \textit{bear-hug-man} is plausible; For \textit{*-snatch-watch}, \textit{girl} is a plausible subject, but not \textit{pig}. Obviously the dexterity of the forefoot of the agent matters here.
\end{enumerate}
The analysis shows that the task and the dataset highlights the necessity for more sophisticated knowledge featurization and cleverer learning techniques (e.g. features from computer vision, propagation methods with stronger capacity to generalize) to reduce the cost of manual annotation.

\section{Conclusion}
\label{sec:conclusion}

We present the novel task of \emph{semantic plausibility}, which forms the foundation of various interesting and complex NLP tasks in event semantics \cite{Bowman:16,mostafazadeh-EtAl:2016:N16-1,Li:17}. We collected a high-quality dedicated dataset, showed empirically that the conventional, distribution data only model fails on the task, and that clever world knowledge injection can help substantially with little annotation cost, which lends initial empirical support for the scalability of our approach in practical applications, i.e. labeling little but propagating well approximates performance with full annotation. Granted that annotation-based injection method does not cover the full spectrum of leverageable world knowledge information (alternative/complementary sources being images and videos, e.g. \citealt{Bagherinezhad:16}), it is indeed irreplaceable in some cases (e.g. features such as \textsc{weight} or \textsc{rigidity} are not easily learnable through visual modality), and in other cases presents a low-cost and effective option. 
Finally, we also discovered the limitation of existing methods through a detailed error analysis, and thereby invite cross-area effort (e.g. multimodal knowledge features) in the future exploration in automated methods for semantic plausibility learning.
\medskip

\section*{Acknowledgments}

This research was supported by NSF grant IIS 1523637. Further, this material is based on research sponsored by DARPA under agreement number FA8750-18- 2-0017. The U.S. Government is authorized to reproduce and distribute reprints for Governmental purposes notwithstanding any copyright notation thereon. We acknowledge the Texas Advanced Computing Center for providing grid resources that contributed to these results. We would also like to thank our reviewers for their insightful comments.

\bibliography{naaclhlt2018}
\bibliographystyle{acl_natbib}

\appendix

\end{document}